\documentclass{article}
\usepackage{ijcai23}

\usepackage{times}
\usepackage{soul}
\usepackage{url}
\usepackage[hidelinks]{hyperref}
\usepackage[utf8]{inputenc}
\usepackage[small]{caption}
\usepackage{graphicx}
\usepackage{amsthm, amssymb}
\usepackage{amsmath}
\usepackage{amsthm}
\usepackage{booktabs}
\usepackage{algorithm}
\usepackage{algorithmic}
\usepackage{xcolor}
\usepackage[switch]{lineno}
\usepackage{multirow}

\theoremstyle{definition}
\newtheorem{definition}{Definition}

\title{A Survey on Intersectional Fairness in Machine Learning: Notions, Mitigation, and Challenges}

\author{
Usman Gohar$^1$ \and
Lu Cheng$^2$ \\
\affiliations
$^1$Iowa State University\\
$^2$University of Illinois Chicago\\
\emails
ugohar@iastate.edu,
lucheng@uic.edu
}

\begin{document}

\maketitle

\begin{abstract}
The widespread adoption of Machine Learning systems, especially in more decision-critical applications such as criminal sentencing and bank loans, has led to increased concerns about fairness implications. Algorithms and metrics have been developed to mitigate and measure these discriminations. More recently, works have identified a more challenging form of bias called intersectional bias, which encompasses multiple sensitive attributes, such as race and gender, together. In this survey, we review the state-of-the-art in intersectional fairness. We present a taxonomy for intersectional notions of fairness and mitigation. Finally, we identify the key challenges and provide researchers with guidelines for future directions.
\end{abstract}
\section{Introduction}

\begin{figure}[]
\includegraphics[width=\columnwidth]{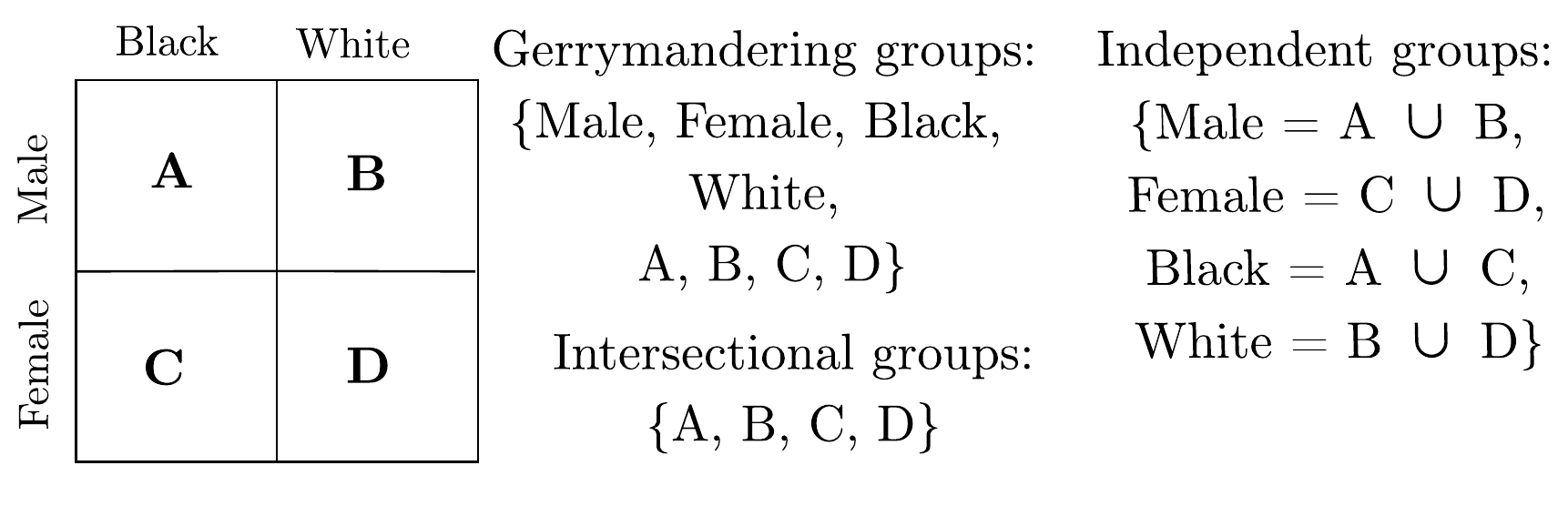}
\caption{Definitions of group fairness~\protect\cite{NEURIPS2020_29c0605a}.}
\label{fig:intersect}
\end{figure}

Machine learning (ML) has been increasingly used in high-stake applications such as loans, criminal sentencing, and hiring decisions with reported fairness implications for different demographic groups \cite{cheng2021socially}. Measuring and mitigating discrimination in ML/AI systems has been studied extensively \cite{10.1145/3457607}. Such works have focused on two specific categories of algorithmic fairness: Group or individual fairness. The majority of early group fairness research was focused on one dimension of group identity, e.g., race or gender. This setting is defined as \textit{independent} groups fairness \cite{NEURIPS2020_29c0605a}. However, recent works have identified a more nuanced case of group unfairness that spans multiple subgroups based on Crenshaw's theory of ``intersectionality'' \cite{Crenshaw1989-CREDTI} called \textit{intersectional group fairness}. At a high level, intersectionality states that interaction along multiple dimensions of identity produces unique and differing levels of discrimination for various possible subgroups, e.g., a Black woman's experience of discrimination differs from both women and Black people in general. Finally, \textit{gerrymandering} groups are the union of independent and intersectional groups. Figure \ref{fig:intersect} shows an example of these group fairness definitions using ``gender'' and ``race''.

By categorizing people \textit{only} into distinct overlapping groups, independent group fairness fails to consider the discrimination people face at the intersection of such groups. This has been well-studied in philosophy and social psychology (e.g., \cite{Bierly1985PrejudiceTC,10.1177/095679761}), but recent works also demand urgency to do so in ML fairness. Specifically, an ML predictor might be fair w.r.t the \textit{independent groups} but not \textit{intersectional groups}. For example, \cite{pmlr-v81-buolamwini18a} identified accuracy disparities that were more significant for Black Women in gender classification algorithms, compared to independent groups. In NLP, works have evaluated popular generative models \cite{Kirk2021BiasOA,10.5555/3454287.3455472} and also identified such cases of intersectional bias. 

Compared to the binary view of fairness in the independent case, the problem of intersectional fairness poses unique challenges. For instance, for what level of granularity of intersectional groups should fairness be guaranteed? On the other hand, smaller subgroups have higher data sparsity, resulting in higher uncertainty \cite{Foulds2018BayesianMO}. Furthermore, an intersectional identity often amplifies biases that might not exist in its constituent groups (e.g., Black woman vs. Black or Woman), rendering traditional mitigation techniques ineffective. To this end, an emerging body of work, e.g., subgroup fairness \cite{pmlr-v80-kearns18a} and multicalibration \cite{pmlr-v80-hebert-johnson18a}, has proposed various notions of intersectional fairness and mitigation techniques that provide a level of guarantee against intersectional discrimination.      

\begin{figure*}[]
\centering
\includegraphics[width=\textwidth]{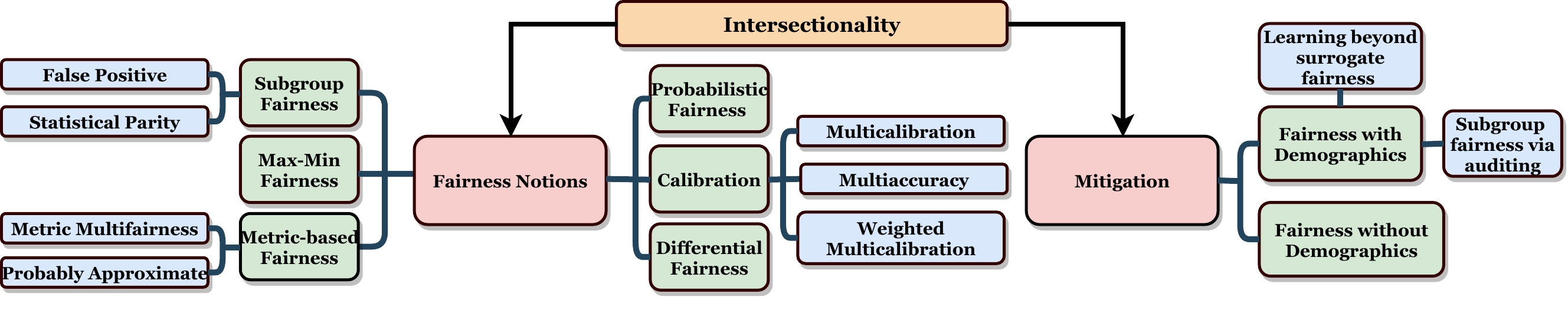}
\caption{The taxonomy for notions of intersectional fairness and fair learning methods.}
\label{fig:taxonomy}
\end{figure*}

Multiple extensive surveys on fairness in ML have been conducted, such as \cite{10.1145/3457607} and \cite{Caton2020FairnessIM}. However, they mainly consider the \textit{independent} group fairness and individual fairness while only briefly discussing \textit{intersectional} cases. To bridge this gap, we review the existing fairness literature on intersectional and gerrymandering groups. In particular, we examine existing notions of intersectional fairness in ML and AI and investigate the techniques that enable fair learning for intersectionality. Our main contributions are:

\begin{enumerate}
  \item We propose the first taxonomy (Fig. \ref{fig:taxonomy}) for the notions of intersectional fairness and fair learning methods for mitigating intersectional discrimination.
  \item We thoroughly examine representative intersectional fairness notions and learning methods and discuss their limitations.
  \item We conclude with the main challenges faced and point out the open problems in the area.
\end{enumerate}
\section{Notions of Intersectional Fairness}
\label{Notions}
Intersectionality, as opposed to group fairness based on independent protected groups (e.g., gender), postulates that the sum of human experiences with discrimination cannot be limited to individual groups alone \cite{Crenshaw1989-CREDTI}. Predictors can appear fair when evaluated on independent groups but not at their intersections \cite{pmlr-v81-buolamwini18a}. Satisfying traditional group fairness for intersectionality is infeasible due to potentially infinite overlapping subgroups. This section reviews fairness notions for intersectionality that limit the number of subgroups by balancing the requirements of group fairness and the stronger notion of individual fairness. \\
\noindent\textbf{Notations}. Each individual is denoted by a tuple $(x, y)$ where $x \in \mathcal{X}$ and $y \in \mathcal{Y}$ denote the instance and ground-truth label, respectively. Let $A = \{s_{1}....s_n\}$ be the set of size $n$ protected attributes, $f$ a predictor, and $f(x)$ the predictor output.

\subsection{Subgroup Fairness}
A pioneering work \cite{pmlr-v80-kearns18a} proposes a stronger notion of group fairness, called \textit{subgroup fairness}, that holds over a large number of \textit{structured} subgroups that can be learned efficiently. In particular, statistical parity (SP) subgroup fairness limits the number of subgroups by disregarding those with limited representations in the data and relaxes the requirement of statistical parity.

Let $\mathcal{C}=\{c: \mathcal{X}\rightarrow\{0,1\}\}$ be a collection of characteristic functions  where $c(s)=1$ indicates that an individual with protected attribute $s$ is in subgroup $c$.

\begin{definition} 
\textit{$f(x)$ is $\gamma-$SP subgroup fair if $\forall$ $c \in \mathcal{C}$:}
{\small
  \begin{multline} \label{stat-parity}
	|P(f(x) = 1) - P(f(x) = 1 | c(s) = 1)|  \\
    \times P(c(s) = 1) \leq \gamma,
   \end{multline}}
\label{eq:subgroup}
\end{definition}

The $\gamma$-SP is determined by the worst-case group $c \in \mathcal{C}$. The first term in Eq. \ref{eq:subgroup} is a penalty on the difference in probability between the positive outcome for a specific subgroup $c$ and for the entire population. The smaller the difference, the fairer the outcome. The second term reweighs the difference by the proportion of the size of each subgroup in relation to the population. Consequently, the unfairness of smaller-sized groups is down-weighted in the final $\gamma-$SP estimation. Thus, it may not adequately protect small subgroups, even if they have high levels of unfairness. Similarly, subgroup fairness can be applied to the false positive (FP) rate.

%\begin{definition} 
%\textit{$f(x)$ is $\gamma-$FP subgroup fair if $\forall c \in \mathcal{C}$:}
%\small
%  \begin{multline} \label{stat-parity}
%	|P(f(x) = 1 | y = 0) - P(f(x) = 1 | c(s) = 1, y = 0)| \\
%    \times P(c(s) = 1, y = 0) \leq \gamma.
%   \end{multline}
%\end{definition}

%This definition is identical to $\gamma$-SP subgroup fairness but excludes subgroups with low negative samples ($y=0$) instead of total samples for each subgroup. 

\subsection{Calibration-based Fairness}

Calibration in binary prediction tasks refers to the accuracy of a predictor's confidence in its predictions \cite{pmlr-v80-kearns18a}. It ensures that the predicted probability distribution for each output class $f(x) = v$ is equal to the actual data probability distribution, i.e., the true expectation is equal to $v$. For example, if six out of ten samples are positive, the underlying probability and expected predicted probability should also be 0.6. Independently, \cite{pmlr-v80-hebert-johnson18a} proposes multicalibration that requires all subgroups to be well-calibrated, assuming access to a class of efficiently-learnable characteristic functions. Formally:

\begin{definition} 
\textit{Given a parameter $\alpha \in [0,1]$, $f(x)$ is ($\mathcal{C}, \alpha$)-multicalibrated if for all predicted values $v \in [0,1]$, $\forall c \in \mathcal{C}$}
{\small
  \begin{equation} \label{multi-calib}
	| \mathbb{E}[c(x)\cdot(y - v)|f(x) = v]| \leq \alpha.
   \end{equation}}
\end{definition}

The parameter $\alpha$ allows for a less stringent requirement on calibration, i.e., a small miscalibration error $\alpha$ is allowed. Intuitively, a rich class $\mathcal{C}$ will contain groups beyond independent cases, such as intersectional groups, leading to stronger fairness guarantees. \textit{Muiltiaccuracy}  \cite{10.5555/3327345.3327393} replaces calibration with accuracy constraints to propose a weaker fairness notion, which requires a predictor to be at least $\alpha$-accurate: $\mathbb{E}[c(x)\cdot f(x) - y(x)] \leq \alpha$ $\forall c \in \mathcal{C}$. Compared to multicalibration, multiaccuracy is less computationally expensive as it is not conditioned on the calibration of each output class across a rich class of intersecting subgroups. %Furthermore, the complexity in multi-class setting is exponential in the number of classes $d$. However, multicalibration provides a stronger fairness gurantee than multiaccuracy. 
These notions define two extremes between efficiency and strong fairness guarantees. To find a balance between the two, \cite{pmlr-v178-gopalan22a} introduces a hierarchy of weighted multicalibration. Formally:

\begin{definition} 
\textit{Given $\mathcal{C}$ and a weight class $\mathcal{W}$, $f(x)$ is $(\mathcal{C},\mathcal{W},\alpha)-$multicalibrated, if $\forall c \in \mathcal{C}$ and $\forall w \in \mathcal{W}$}
{\small
  \begin{equation} 
  \label{multi-calib}
	| \mathbb{E} [c(x). w(f(x))(y - f(x)]| \leq \alpha.
   \end{equation}}
\end{definition}

\begin{figure}[]
\centering
\includegraphics[scale = 0.9]{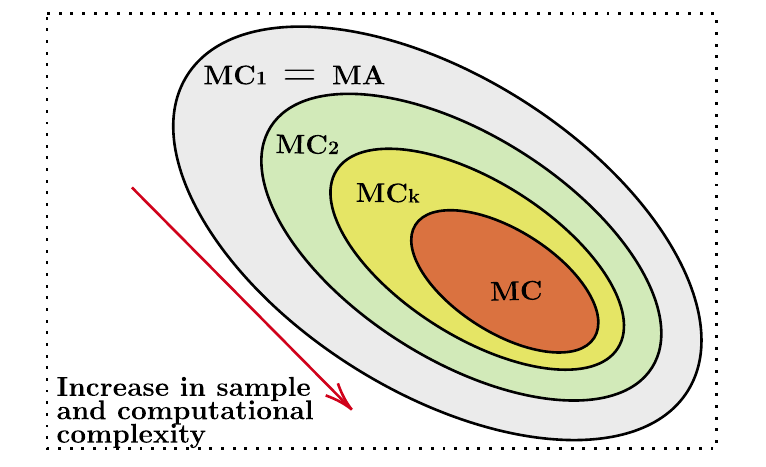}
\caption{Hierarchy of multicalibration that interpolates from multiaccuracy (MA) to mutilcalibration (MC)~\protect\cite{pmlr-v178-gopalan22a}.}
\label{fig:hierarchy}
\end{figure}

The choice of class $\mathcal{W}$ can lead to multiple variations of the multicalibration notions. \textit{Low-degree multicalibration} is defined as taking the weight function $w(f(x))$ to be a class of $k-1$ polynomials with $k$ degrees. As shown in Figure \ref{fig:hierarchy}, when $k = 1$, $w(f(x))$ is constant, and we get the efficient albeit weaker fairness notion of \textit{multiaccuracy}. At higher degrees of polynomial, it converges to \textit{multicalibration}. For a class of 1-Lipstichz functions, we get the ($C, \alpha)$-\textit{smooth-multicalibration}. Finally, if the predictor is calibrated on prediction intervals instead of calibrating on each predicted value, we arrive at \textit{full-multicalibration}. As such, this hierarchy interpolates the space between multiaccuracy and multicalibration, increasing the strength of fairness guarantees and complexity at higher levels.

\subsection{Metric-based Fairness}

Another line of work \cite{yona2018probably} addressed the computational concerns of satisfying fairness for possibly large number of intersectional groups by relaxing the notion of the seminal work of \cite{10.1145/2090236.2090255} on individual fairness. Individual fairness requires that given a similarity metric if the distance between a pair of individuals is small, a predictor should output similar classification distributions. Inspired by this, \cite{yona2018probably} proposes a relaxed generalization of individual fairness which allows a small fairness error, called \textit{approximate-metric} fairness. Similar to subgroup fairness and multicalibration works, the relaxation allows the use of efficient learning algorithms that protects every sufficiently-large subgroup. However, unlike those works, the subgroups are not defined a priori. Formally:

\begin{definition} 
\textit{For a small $\alpha \in [0,1]$ and $\gamma \in [0,1]$, $f(x)$ is ($\alpha,\gamma)$-approximately metric fair w.r.t similarity metric $d$ and data distribution $\mathcal{D}$ if }
  \begin{equation} \label{approximate 
  metric-multifairness}
  \small
	\underset{(x, x^{'})\sim \mathcal{D}}{\mathbb{P}}[|f(x) - f(x^{'})| \geq d(x,x^{'}) + \gamma] \leq \alpha,
  \end{equation}
where $(x,x^{'})$ are two individuals sampled from the dataset.
\end{definition} 

The parameters $\alpha$ and $\gamma$ allow for small errors in similarity and the metric-fairness measures. Approximate-metric fairness requires that individual fairness holds for all but a small fraction of $\alpha$ pairs of individuals. Consequently, it protects all subgroups of size greater than $\alpha$ as members within the subgroups are treated similarly to those outside. Approximate-metric fairness assumes that the similarity metric is already known for individuals. To relax the assumption, \cite{10.5555/3327345.3327393} introduces \textit{metric-multifairness}, which supports any similarity metric and requires that similar \textit{subgroups} are treated similarly based on the average distance between individuals in those groups. Formally:

\begin{definition} 
\textit{For a small constant $\gamma > 0 $ and an unknown similarity metric $d$, $f(x)$ is $(C,d, \tau)$-metric mutlifair if }
{\small
  \begin{equation} 
  \label{metric-multifairness}
	\underset{(x, x^{'})\sim A}{\mathbb{E}}[|f(x) - f(x^{'})|] \leq \underset{(x, x^{'})\sim A}{\mathbb{E}} [d(x,x^{'})] + \gamma.
  \end{equation}}
\end{definition}

More specifically, it requires that individuals in subgroups are treated differently only if they differ substantially from the average difference between individuals within the subgroup. There exist other works (e.g., \cite{10.5555/3327144.3327185}) that define metric-based fairness in online learning; however, we limit ourselves to the offline setting due to space constraints.

\subsection{Differential Fairness}

Anti-discrimination laws \cite{commission1978guidelines} in the United States declare an outcome as biased if the ratio of probabilities of a favorable outcome between an advantaged and disadvantaged group is less than 0.8. \textit{Differential Fairness} ({DF}) \cite{9101635} extends this rule to protect multidimensional intersectional categories. But instead of using a fixed threshold at 80\%, DF used a sliding scale, similar to the concept of ``differential privacy" \cite{10.1007/11787006_1}, to measure the unfairness of a predictor w.r.t intersectional groups.

\begin{definition} 
\textit{$f(x)$ is $\epsilon-$differentially fair if}
{\small
  \begin{equation} \label{diff-fair}
	e^{-\epsilon} \leq \frac{P(f(x) = y| s_i)}{P(f(x) = y| s_j)} \leq e^\epsilon,
  \end{equation}}
  holds for all tuples $(s_i, s_j) \in A \times A$ where $0\leq P(s_j) \leq 1.$
\end{definition}

For small values of $\epsilon$, the DF criterion states that probabilities of favored outcomes will be similar for any combination of intersectional groups. Unlike other notions, DF ensures fairness for all possible groups, regardless of their size. To estimate the probabilities in Eq. \ref{diff-fair}, empirical counts for each subgroup can be used. However, it suffers from data sparsity at higher intersections of groups. This can be addressed by using a Dirichlet prior. Finally, \cite{9101635} also proposes \textit{DF-bias amplification}, which measures the discrimination of a predictor by taking the difference of the DF of the dataset ($\epsilon_1$) and the predictor ($\epsilon_2$).

Furthermore, \cite{DBLP:journals/corr/abs-1911-01468} extended the DF notion to other standard group fairness notions such as Statistical Parity, Equality of Opportunity, False Positive Rate Parity, and equalized odds. The ratio in Eq. \ref{diff-fair} is replaced with the specific group fairness definition for which the $\epsilon$ is measured.

\subsection{Max-Min Fairness}
The \textit{Max-min} (or \textit{min-max}) notion of fairness is based on the Rawlsian principle of distributive justice \cite{rawls2001justice}. This principle allows for inequalities but aims to maximize the minimum utility across different protected groups. Given a predictor and a fairness metric, it aims to maximize the fairness of the worst-off subgroup. The Max-Min Fairness \cite{Ghosh2021CharacterizingIG}  is extended to intersectional cases by measuring the fairness of any combination of intersectional subgroups using existing fairness definitions and then taking the ratio of the maximum and minimum values from this list of subgroups. A ratio below 1 indicates a disparity between groups, with greater disparity if the ratio is closer to 0. This ratio can be applied to any existing fairness or performance measures like AUC. However, this definition also suffers from low data sparsity when the number of dimensions of intersectionality increases.

\subsection{Probabilistic Fairness}

Differential fairness uses a Dirichlet prior of uniform parameter $\alpha$ to resolve the issue of intersectional groups having zero counts in the data. The parameter affects this empirical count approach and may miss high-risk subgroups not represented in the data. To solve this, \textit{Probabilistic Fairness} \cite{molina2022bounding} relaxes the requirement of guaranteeing fairness for all subgroups using a probabilistic approach. Formally,

\begin{definition} 
\textit{For $\epsilon \geq 0$ and $\delta \in [0,1]$, a predictor is $(\epsilon,\delta)$-probably intersectionally fair if}
{\small
  \begin{equation} \label{prob-fair}
	P(U \geq \epsilon) \leq \delta,
  \end{equation}}
where $U = u(f(x),s,s^{'})$ measures unfairness for a randomly chosen prediction and two protected groups ($s \neq s^{'}$) to compare them.  Probabilistic fairness captures the expected size $\delta$ of the population for which the predictor discriminates more than $\epsilon$.
\end{definition}

\subsection{Discussions}

In contrast to traditional fairness metrics, intersectional fairness notions encapsulate a greater number of more granular subgroups and intersectional identities. Existing notions of intersectional fairness depend on the level of unfairness experienced by the most disadvantaged group across all intersections of individual groups. These notions only differ in their approaches for identifying and limiting the vast number of such subgroups to efficiently measure fairness. One risk with designing intersectional notions is that they arbitrarily limit subgroups based on various methods that can efficiently identify them, hence, participating in the same fairness gerrymandering it attempts to solve \cite{10.1145/3531146.3533114}. For instance, subgroup fairness uses a weight term to prove generalization guarantees w.r.t the underlying population; however, by doing so, it down-weights minority groups which fails to adequately protect them. This highlights the need for a broader involvement of stakeholders to design notions and not simply rely on computational methods. While Max-Min and Differential Fairness truly encapsulates all such groups without disregarding any subgroup, they suffer from data sparsity. Probabilistic approaches might be favorable in such scenarios. Finally, future work could explore the applicability of these definitions in other domains (e.g., recommender systems, NLP, and so on) and define notions for continuous attributes.

\section{Improving Intersectional Fairness}

ML systems have been shown to exhibit unfairness due to biases in data \cite{10.1145/3457607} and algorithms \cite{pmlr-v81-buolamwini18a,gohar2022towards}. In response, there have been great efforts to mitigate bias and improve fairness in ML systems. However, comparatively fewer fair learning algorithms were proposed to address the unique challenges of intersectional fairness. Here, we review the two lines of approaches for intersectional fairness learning: \textit{Intersectional Fairness with Demographics} and \textit{Intersectional Fairness Without Demographics}.

\subsection{Intersectional Fairness with Demographics}

A surge of methods have been proposed to mitigate bias by learning fair models \cite{10.1145/3457607}. These techniques are generally applied to the training data (pre-processing), the learning algorithm (in-processing), or the predictions (post-processing). Intersectional fairness with demographics mostly falls into the last two categories using the specific intersectional fairness notions we discussed in Section \ref{Notions}.

\paragraph{Subgroup fairness via auditing.} A number of works \cite{pmlr-v80-kearns18a,10.5555/3327345.3327393,10.1145/3306618.3314287,pmlr-v80-hebert-johnson18a} use auditing to learn fair predictors w.r.t a large number of subgroups. This approach involves an auditor, with access to i.i.d. samples $X$ from an unknown distribution, that assesses the fairness of a predictor using a fairness metric and identifies subgroups with high unfairness. Then a learning algorithm tries to minimize error subject to that fairness constraint. Separately, \cite{pmlr-v80-hebert-johnson18a,pmlr-v80-kearns18a} both prove that the task of learning such a fair model is equivalent to auditing an arbitrary predictor w.r.t a class of subgroups $\mathcal {C} $ which is computationally equivalent to weak agnostic learning of $\mathcal{C}$. Utilizing this approach, the seminal work of \cite{pmlr-v80-kearns18a} proposes a zero-sum game between an \textit{Auditor} and a \textit{Learner} for the subgroup fairness notion. In this setting, the zero-sum game is a Fictitious Play using a cost-sensitive classification oracle \cite{agarwal2018a}. Instead of auditing during training, \cite{pmlr-v80-hebert-johnson18a} use a post-processing iterative boosting algorithm by combining all $c \in \mathcal{C}$ until the model is $\alpha-$calibrated. Multiaccuracy \cite{10.1145/3306618.3314287} extends this approach to learning a multi-accurate predictor that guarantees accurate predictions w.r.t $\mathcal{C}$. 
%Additionally, they ensure that the initial model is well-calibrated for an identifiable subgroup, the post-processed model will be too. 
Inspired by these approaches, \cite{10.5555/3327345.3327393} propose a variant of stochastic descent gradient that can be leveraged using auditing to post-process a predictor. 

\paragraph{Learning beyond surrogate fairness notions.} Techniques discussed so far are strictly based on surrogate fairness notions that are adapted to apply to many subgroups e.g., SP-subgroup fairness is a surrogate of the statistical parity notion. Next, we discuss works that go beyond tailor-made intersectional fairness notions. The approach outlined in \cite{shui2022on} focuses on addressing group sufficiency for many subgroups (including intersectional groups) in ML predictors. Group sufficiency states that given the prediction $f(x)$, the conditional expectation of the ground-truth label ($\mathbb{E}[Y|(f(x), A]$) is similar across different subgroups. They first derive an upper bound of the group sufficiency gap and propose a bi-level optimization approach with a randomized algorithm that generates the output distribution of the predictor. In the lower level, subgroup-specific output distribution is learned using a small sample of each subgroup's labeled data. Then, the final output distribution is updated at the upper level to ensure it is close to all subgroup-specific output distributions. Another work called \textit{GroupFair} \cite{NEURIPS2020_29c0605a} proposes Bayes-optimal predictors that are fair for all subgroups w.r.t loss, using a weighted Empirical Risk Minimization (ERM) oracle \cite{agarwal2018a}. Recently, \cite{kang2022infofair} took another step towards a more generalized mitigation approach that does not depend on self-defined fairness notions by capturing linear and non-linear dependence between predictions and intersectional groups using mutual information \cite{shannon1948mathematical}. Finally, \cite{DBLP:journals/corr/abs-1911-01468}, combines and extends previous works \cite{NIPS2016_9d268236,10.1145/3097983.3098095} to include intersectional cases, by utilizing differential fairness. This involves randomly flipping predictions and a loss function which allows users to find the optimal fairness-accuracy trade-off.

\subsection{Intersectional Fairness without Demographics}

This line of research addresses intersectional biases without using protected attribute information due to privacy laws. Additionally, data sparsity in smaller subgroups and normative concerns about using synthetic data generation techniques \cite{10.1145/3531146.3533101} make this a compelling approach to tackle intersectional biases. It exploits the correlations between protected attributes and non-protected attributes to approximate the subgroup information. Most of the existing works are in-processing methods.

One such line of work \cite{Hashimoto2018FairnessWD,NEURIPS2020_07fc15c9,pmlr-v139-martinez21a} aims to maximize the minimum utility for all subgroups by using Rawlsian's Max-Min theory. Unlike parity-based fairness notions, this principle argues in favor of reducing worst-case risk. The fairness objective to minimize the worst-case loss can be formulated as:
{\small
\begin{equation}
    L_{\max}(f) = \underset{s \in S}{\max}  
 \mathbb{E}[l(f;X)].
\end{equation}}

\noindent where $\mathbb{E}[l(f;X)]$ is the expected loss for a loss function e.g., log loss. One approach \cite{Hashimoto2018FairnessWD} uses \textit{distributionally robust optimization} (DRO) to minimize the worst case loss for any subgroup. DRO achieves this by minimizing the loss over all distributions that are close to the input distribution. Their approach considers the worst-case loss over all distributions with $\chi^2-$divergence less than $r$, where $r$ is the radius of a chi-squared ball ($B(P,r)$) around the input probability distribution $P$. The DRO function is defined as: 

{\small
 \begin{equation}
L_{DRO}(f,r) = \underset{Q \in B(P,r)}{sup} \mathbb{E_Q}[l(f;X)].
 \end{equation}}
 
Specifically, DRO attempts to reduce the possibly exponential number of subgroups by only considering worst-case distributions that exceed a given size $\alpha$. A key distinction here is that the objective of the learning algorithm does not depend on $\alpha$. Consequently, all subgroups have equal representation in the loss function to be minimized. However, this method can potentially optimize noisy outliers, reducing its effectiveness. 

To address the limitations of DRO, \cite{NEURIPS2020_07fc15c9} propose an Adversarial Reweighting-based approach that relies on the notion of \textit{computationally-identifiable} groups \cite{pmlr-v80-hebert-johnson18a}. They design a minimax game between a \textit{learner} and \textit{adversary}: the \textit{learner} is trained to minimize the expected loss while an adversarial neural network is tasked to learn identifiable regions where the learner has significant errors. Their results show that the regions with high errors correspond to various intersectional groups such as \textit{black-female}. The other recent method \cite{pmlr-v139-martinez21a} based on the Max-Min objective proposes a Pareto-efficient \cite{mas-colell.whinston.ea95} learning algorithm to provide a performance guarantee for unidentified protected class w.r.t. to user-defined minimum group size. Although these works target minimizing the worst-case performance of \textit{any} unknown subgroup of a minimum size, experimental results show that they also improve fairness for some intersectional groups.

\subsection{Discussion}

One of the main drawbacks of current works is that the majority of them rely on specific surrogate fairness notions that we discussed in Section \ref{Notions}. Furthermore, these works ignore certain subgroups that do not conform to specific statistical requirements, e.g., computationally identifiable, that reinforces fairness gerrymandering. One approach to tackle this can be to learn latent representations of intersectional groups that can be then de-biased using geometric approaches \cite{cheng2022debiasing}. 
Intersectional fairness without demographics is a promising direction to mitigate intersectional bias, but the current works are limited to the Max-Min notion. While there are certain applications (e.g., healthcare) where improving the utility of the worst-case groups is an important goal, many other applications can be required by law to ensure parity for all subgroups. Similarly, it is critical to evaluate the effectiveness of these methods on different intersectional groups present in the data. For instance, \cite{NEURIPS2020_07fc15c9} relies on computational identifiability, which depends on correlations with unprotected attributes. Such methods might fail for intersectional groups that do not have strong demographic signals present in unprotected attributes, hence, failing to protect such groups. Future research can explore learning predictive patterns for underrepresented intersectional groups by leveraging common patterns shared with related groups. For example, Black Females and Black Males might have common structural patterns \cite{10.1145/3531146.3533101}.

%these works are only limited to the min-max fairness notions. More work is needed to develop fair learning algorithms for other fairness notions and applications. Also there is a difference between subgroup and intersectional fairness. CLARIFY

%Another negative is that these ideas will not work for user-defined intersectional groups or subgroups that are missing or too small. There needs to be a debate on how small of a group to define

\section{Applications}
Most works discussed above are generally focused on classification tasks with i.i.d data. In this section, we review the application of intersectional fairness in other domains of AI.
\subsection{Natural Language Processing}
Numerous works (e.g. \cite{10.5555/3454287.3455472}) have observed that the societal bias inherent in real-world corpora translates to discrimination in NLP models. More recently, there has been a greater effort to focus on benchmarking and debiasing NLP models along intersectional lines.

\paragraph{Benchmark.} Several studies have examined bias in sentiment analysis systems, such as \cite{kiritchenko2018examining}, and found that such systems discriminate based on intersections of gender and race. A closely related study by \cite{camara-etal-2022-mapping} across multiple languages confirms these biases. Contextualized word embedding models, including GPT-2 and BERT, have been analyzed for gender and race intersections at sentence level \cite{may2019measuring} and at contextualized word level \cite{10.5555/3454287.3455472}. These works report higher discrimination at the intersection of race and gender (e.g., Black females) compared to either group alone. Separately, \cite{hassan-etal-2021-unpacking-interdependent} evaluates BERT for discrimination against people with disabilities along similar intersectional groups. To automatically identify intersectional biases in static word embeddings, \cite{10.1145/3461702.3462536} introduces Contextualized Embedding Association Test (CEAT) to measure intersectional bias in contextualized settings. Finally, \cite{Kirk2021BiasOA} expands upon these works to incorporate intersections of religion, sexuality, and political affiliations to investigate representational and allocational harms concerning occupational stereotypes in language models. To quantify the scope of the intersectional bias problem in NLP, \cite{lalor-etal-2022-benchmarking} performs a comprehensive evaluation of state-of-the-art NLP models and debiasing strategies for intersectional bias, benchmarking ten downstream tasks and five demographic groups. These studies highlight the importance of considering a diverse set of intersecting groups in discussions around bias in language models, especially user-facing large language models.
\paragraph{Mitigation.} Relatively few works have focused on debiasing along intersectional dimensions. The earliest work by \cite{subramanian-etal-2021-evaluating} evaluates two debiasing techniques and shows that debiasing methods based on independent groups are prone to gerrymandering. To address the issue of limited data for intersectional groups, \cite{cheng2022debiasing} introduces JoSEC, a debiasing approach that leverages the nonlinear geometry of subspace representations to learn intersectional subspace without using predefined word sets. Unlike the linear correlation assumption, they posit that the individual subspaces intersect over a single dimension where the intersectional group subspace resides. 
\subsection{Ranking Systems}
Another common application domain is ranking systems. Fair ranking refers to the method of ensuring that ranking and recommender systems are equitable for all parties involved, including users, providers, and the items being ranked \cite{10.1145/3457607}. Here we review such works that explore the problem through an intersectional lens. 

\paragraph{Top-$K$ Ranking.} In the context of fair top-$k$ selection, \cite{Barnab2020IntersectionalAA} examines discrimination along twelve intersections of socioeconomic status, high-school type, and zip code regions for college admissions and proposes an algorithm to select candidates with high utility whilst giving more representation to disadvantaged intersectional groups. Another promising approach \cite{yang2021causal} uses a causal framework for fair ranking across intersections of gender and race. They compute model-based counterfactuals and rank the resulting scores accordingly. Counterfactual fairness denotes that a prediction is fair if the outcome of an AI system does not change when a single variable is changed and all else remains the same \cite{kusner2017counterfactual}.  

\paragraph{Fair Rank Aggregation.} A similar problem of fair rank aggregation requires combining various rankings to create a consensus ranking, but this can be biased against individual protected attributes like gender, race, and intersectional groups. To resolve this, \cite{Cachel2022MANIRankMA} proposes a group fairness criterion for consensus ranking that ensures fairness for individual groups and their intersections. The unified fairness notion ensures minimal statistical parity difference between pairs of candidate rankings for individual and intersectional groups: $ARP_{a_{k}} \leq \Delta  (\forall a_{k} \in A$ and $IRP \leq \Delta $, where $\Delta$ represents desired closeness to statistical parity (zero ensures parity), $ARP$ and $IRP$ represent rank parity for individual and intersectional groups, respectively. They use empirical counts to measure unfairness for each group using this metric, which is prone to data sparsity. 
\subsection{Auditing and Visualization}
Auditing evaluates the fairness of AI systems after training a predictor. It is useful to detect discrimination against a large number of possibly intersecting subgroups. Auditing can identify such subpopulations and make the model more transparent by highlighting its failures. One such work \cite{10.1145/3306618.3314287} leverages a decision-tree-based auditing model to identify bias against dark-skinned women in an image dataset. Other works such as \cite{pmlr-v80-hebert-johnson18a} and \cite{pmlr-v80-kearns18a} utilize this approach  to train fair predictors w.r.t a large number of subgroups.

Some works have created visualization tools to detect potentially discriminatory data subsets. FairVis, \cite{Cabrera2019FAIRVISVA}, is a visual analytic tool for experts to utilize their domain knowledge in generating subgroups and augmenting automated detection and mitigation strategies. The tool uses clustering analysis to identify statistically similar subgroups and then computes important features and fairness metrics using entropy.  Another such tool \cite{10.1145/3411763.3451587} identifies intersectional biases encoded in word embeddings. Given a pre-trained word embedding, it computes a bias score (using cosine distance) for each subgroup (e.g., male/female for a binary gender) and predefined word sets. A discriminatory word is considered to be associated with an intersectional group if it strongly associates with each of its individual groups according to the bias score. However, this approach may overlook cases like ``Hair Weaves'', which are associated with intersectional groups (Black Female) but not individual subgroups (Black or Female).

A software engineering approach \cite{10.1145/3318464.3384689} seeks to ensure adequate representation for relevant intersectional groups within the dataset using coverage. They define intersectional groups using patterns, e.g., \{Gender=Female, Race = Black\}. Then it requires that intersectional subgroups have a minimum threshold $\tau$ of instances. Finally, \cite{8713886} finds intersectional bias in the dataset by dividing it into more granular groups until a subgroup with significant loss is found. 
\subsection{Discussion}
Current applications of intersectionality have been focused on NLP and Ranking. Some other promising applications include recommender systems \cite{islam2019mitigating}, graph embeddings \cite{bose2019compositional}, computer vision \cite{10.1145/3442188.3445932}, and so on. It is imperative that existing ML systems are holistically evaluated under the intersectional framework to aid in developing inclusive and fair ML systems. In NLP, a limitation of current works is the assumption that demographic information is available. Given increasing regulatory and privacy concerns, more research is needed to understand potential correlations in data that can be leveraged to tackle intersectional biases.

\section{Datasets and Evaluation Metrics}

% \usepackage{multirow}
\begin{table}[]
\centering
\small
\renewcommand{\tabcolsep}{3.5pt}
%\resizebox{\columnwidth}{!}{
\begin{tabular}{lll}
\toprule
\textbf{Type} & \textbf{Dataset}                                                  & \textbf{Demographics}                          \\ \toprule
\multirow{4}{*}{Tabular} & Adult                          & Gender, Age, Race                                            \\ 
   &Student                                                        & Gender, Age, Alcohol, Relationship                           \\                         
& Law School                                                     & Gender, Age, Race, Income                                    \\ 
& Compass                                                        & Gender, Race                                              \\  \hline
\multirow{4}{*}{NLP} & Psychometrics                         & Gender, Age, Race, Income, Education                         \\ 
& MTC                                  & Gender, Age, Race                                           \\  
& FIPI                                 & Gender, Age, Race, Income, Education                        \\ 
& MBTI                                 & Gender, Age                                                  \\ \hline
\multirow{2}{*}{Ranking} & MEPS & Gender, Race, Age \\  
& MovieLens & Gender, Age, Occupation \\ \hline 

\multirow{3}{*}{Image} & CelebA & Gender, Age, Race \\
& UTKFace & Gender, Age, Ethnicity \\
& PPB & Gender, Race  \\
\bottomrule
\end{tabular}%}
\caption{Summary of popular datasets across different AI domains that contain multiple intersectional groups. Law School ~\protect\cite{Wightman1998LSACNL} and Compass ~\protect\cite{angwin_larson_kirchner_mattu_2016} are also used in Ranking. Adult ~\protect\cite{Dua:2019}, Student ~\protect\cite{Cortez2008UsingDM}, Psychometrics ~\protect\cite{abbasi-etal-2021-constructing}, Multilingual Twitter Corpus (MTC) ~\protect\cite{huang-etal-2020-multilingual}, Five Item Personality Inventory (FIPI) and Myers-Briggs Type Indicator (MBTI) ~\protect\cite{gjurkovic-etal-2021-pandora}, MovieLens ~\protect\cite{10.1145/2827872}, MEPS ~\protect\cite{rankdata}, CelebA ~\protect\cite{liu2018large}, UTKFace ~\protect\cite{zhifei2017cvpr}, PPB ~\protect\cite{pmlr-v81-buolamwini18a}.}
\label{tab:datasets}
\end{table}

\paragraph{Available Datasets.} Data scarcity is a big challenge for intersectional fairness as the number of dimensions increase. In Table \ref{tab:datasets}, we summarize some popular datasets with adequate intersectional groups, across different AI domains. We hope our consolidated summary provides researchers with convenient access to datasets with rich subgroup information.

\paragraph{Evaluation Metrics.} Most studies use the intersectional notions they define for the learning algorithm as the evaluation metrics. Beyond that, worst-case classification accuracy and AUC are broadly used when demographic information is unavailable. These worst-case metrics have also been adopted in NLP and Image classification tasks.

\section{Summary and Open Problems}

In this survey, we review recent advances in the fairness of ML systems from an intersectional perspective. Intersectionality poses unique challenges that traditional bias mitigation algorithms and metrics cannot effectively address. We review different definitions of group fairness, present a taxonomy of intersectional fairness notions and mitigation methods, and review the literature on intersectionality in other AI domains. Next, we briefly discuss open problems and potential future research directions.

\paragraph{Data Sparsity.}  The lack of representative data for marginalized subgroups is a significant challenge. Alternative approaches that do not rely on demographic information may be employed, but these methods do not guarantee that bias against missing subgroups will be addressed. Therefore, a concerted effort to create more inclusive datasets is needed. 

\paragraph{Selecting subgroups.} Most works propose fairness notions that guarantee fairness only for a limited number of subgroups that are considered statistically meaningful (computationally feasible). This approach fails to protect minority subgroups that do not conform to these statistical requirements. Relying solely on such computational methods reinforces fairness gerrymandering \cite{10.1145/3531146.3533114}. Hence, it is crucial to involve diverse stakeholders to ensure that the needs and perspectives of different intersectional groups are met. 

\paragraph{Generalized mitigation approaches.} Existing works on mitigating intersectional bias propose learning algorithms based on specific surrogate fairness notions. These cannot be generalized to other predictors to be used as plug-in mitigation tools. Learning latent representations for intersectional groups so debiased data can be used with any predictor and for any classification task, is a potential direction.

%Another area to explore is to generate test cases to audit predictors for intersectional biases. Previous works have used techniques such as search-based methods to find test cases where predictors are unfair. With the added complexity of intersectionality, it would be beneficial to design computationally-feasible search algorithms to generate test cases to audit user-facing models for intersectional bias.\\ %Domain expertise can be 
%incorporated to prevent discrimination against certain subgroups, such as granular Asian groups like Indian and Hmong.

\paragraph{Intersectional fairness beyond parity.} Current research overlooks the under-representation of intersectional groups by solely focusing on achieving parity \cite{10.1145/3531146.3533114}. While it is useful, unequal distribution may be fairer in certain cases. For instance, equalizing hiring rates cannot fix the under-representation of Black females in tech. Therefore, more research on non-distributive intersectional fairness is needed.

\paragraph{Generating test cases for auditing.} Generating test cases to audit predictors for intersectional biases is another important direction. With the added complexity of intersectionality, it would be beneficial to evaluate previous testing tools \cite{Chen2022FairnessTA} and design new tools to test user-facing models for intersectional bias. This can help identify intersectional subgroups against which the predictor is discriminatory. 

\paragraph{Beyond mitigation.} To effectively address intersectional bias in ML systems, it's crucial to understand its propagation throughout the ML development cycle, from data collection to algorithms. Exploring causal approaches to understanding intersectional bias is one such interesting direction. 

%Another area to explore is developing methods to generate test cases to audit predictors for intersectional biases. Previous works \cite{Chen2022FairnessTA} have used techniques such as search-based methods to find test cases where predictors are unfair. With the added complexity of intersectionality, it would be beneficial to design computationally-feasible search algorithms to generate test cases to audit user-facing models for intersectional bias. \\

\paragraph{Evaluating fairness notions.} Intersectional notions proposed for handling biases have mostly been explored theoretically, with little evidence of their effectiveness on real-world datasets, especially in evaluating the subgroups they fail to protect. Though there are no simple solutions for dealing with intersectional biases in ML, we must measure and benchmark such biases to tackle this problem effectively.

\section*{Acknowledgements}
This material is based upon work supported by the Cisco Research Gift Grant.

%\section*{Ethical Statement}

%There are no ethical issues.

%\section*{Acknowledgments}

{\small
\bibliographystyle{named}
\bibliography{ijcai23}}

\begin{thebibliography}{}

\bibitem[\protect\citeauthoryear{Abbasi \bgroup \em et al.\egroup
  }{2021}]{abbasi-etal-2021-constructing}
Ahmed Abbasi, David Dobolyi, John~P. Lalor, Richard~G. Netemeyer, Kendall
  Smith, and Yi~Yang.
\newblock Constructing a psychometric testbed for fair natural language
  processing.
\newblock In {\em EMNLP}, pages 3748--3758. ACL, 2021.

\bibitem[\protect\citeauthoryear{Agarwal \bgroup \em et al.\egroup
  }{2018}]{agarwal2018a}
Alekh Agarwal, Alina Beygelzimer, Miroslav Dud{\'\i}k, John Langford, and Hanna
  Wallach.
\newblock A reductions approach to fair classification.
\newblock In {\em ICML}, pages 60--69. PMLR, 2018.

\bibitem[\protect\citeauthoryear{Akrami \bgroup \em et al.\egroup
  }{2011}]{10.1177/095679761}
Nazar Akrami, Bo~Ekehammar, and Robin Bergh.
\newblock Generalized prejudice: Common and specific components.
\newblock {\em Psychological Science}, 22(1):57--59, 2011.

\bibitem[\protect\citeauthoryear{Angwin \bgroup \em et al.\egroup
  }{}]{angwin_larson_kirchner_mattu_2016}
Julia Angwin, Jeff Larson, Surya Mattu, and Lauren Kirchner.
\newblock Machine bias.
\newblock In {\em Ethics of data and analytics}, pages 254--264. Auerbach
  Publications.

\bibitem[\protect\citeauthoryear{Barnab{\`o} \bgroup \em et al.\egroup
  }{2020}]{Barnab2020IntersectionalAA}
Giorgio Barnab{\`o}, Carlos Castillo, Michael Mathioudakis, and Sergio Celis.
\newblock Intersectional affirmative action policies for top-k candidates
  selection.
\newblock {\em ArXiv}, abs/2007.14775, 2020.

\bibitem[\protect\citeauthoryear{Bierly}{1985}]{Bierly1985PrejudiceTC}
Margaret~M. Bierly.
\newblock Prejudice toward contemporary outgroups as a generalized attitude.
\newblock {\em Journal of Applied Social Psychology}, 15:189--199, 1985.

\bibitem[\protect\citeauthoryear{Bose and
  Hamilton}{2019}]{bose2019compositional}
Avishek Bose and William Hamilton.
\newblock Compositional fairness constraints for graph embeddings.
\newblock In {\em ICML}, pages 715--724. PMLR, 2019.

\bibitem[\protect\citeauthoryear{Buolamwini and
  Gebru}{2018}]{pmlr-v81-buolamwini18a}
Joy Buolamwini and Timnit Gebru.
\newblock Gender shades: Intersectional accuracy disparities in commercial
  gender classification.
\newblock In {\em In FAT}, pages 77--91. PMLR, 2018.

\bibitem[\protect\citeauthoryear{Cabrera \bgroup \em et al.\egroup
  }{2019}]{Cabrera2019FAIRVISVA}
{\'A}ngel~Alexander Cabrera, Will Epperson, Fred Hohman, Minsuk Kahng, Jamie~H.
  Morgenstern, and Duen~Horng Chau.
\newblock Fairvis: Visual analytics for discovering intersectional bias in
  machine learning.
\newblock {\em IEEE VAST}, 2019.

\bibitem[\protect\citeauthoryear{Cachel \bgroup \em et al.\egroup
  }{2022}]{Cachel2022MANIRankMA}
Kathleen Cachel, Elke~A. Rundensteiner, and Lane Harrison.
\newblock Mani-rank: Multiple attribute and intersectional group fairness for
  consensus ranking.
\newblock {\em IEEE ICDE}, 2022.

\bibitem[\protect\citeauthoryear{C{\^a}mara \bgroup \em et al.\egroup
  }{2022}]{camara-etal-2022-mapping}
Ant{\'o}nio C{\^a}mara, Nina Taneja, Tamjeed Azad, Emily Allaway, and Richard
  Zemel.
\newblock Mapping the multilingual margins: Intersectional biases of sentiment
  analysis systems in {E}nglish, {S}panish, and {A}rabic.
\newblock In {\em LT-EDI}. ACL, 2022.

\bibitem[\protect\citeauthoryear{Caton and Haas}{2020}]{Caton2020FairnessIM}
Simon Caton and Christian Haas.
\newblock Fairness in machine learning: A survey.
\newblock {\em ArXiv}, abs/2010.04053, 2020.

\bibitem[\protect\citeauthoryear{Chen \bgroup \em et al.\egroup
  }{2022}]{Chen2022FairnessTA}
Zhenpeng Chen, J~Zhang, Max Hort, Federica Sarro, and Mark Harman.
\newblock Fairness testing: A comprehensive survey and analysis of trends.
\newblock {\em ArXiv}, abs/2207.10223, 2022.

\bibitem[\protect\citeauthoryear{Cheng \bgroup \em et al.\egroup
  }{2021}]{cheng2021socially}
Lu~Cheng, Kush~R Varshney, and Huan Liu.
\newblock Socially responsible ai algorithms: Issues, purposes, and challenges.
\newblock {\em JAIR}, 71:1137--1181, 2021.

\bibitem[\protect\citeauthoryear{Cheng \bgroup \em et al.\egroup
  }{2022}]{cheng2022debiasing}
Lu~Cheng, Nayoung Kim, and Huan Liu.
\newblock Debiasing word embeddings with nonlinear geometry.
\newblock In {\em COLING}, pages 1286--1298, 2022.

\bibitem[\protect\citeauthoryear{Chung \bgroup \em et al.\egroup
  }{2020}]{8713886}
Y.~Chung, T.~Kraska, N.~Polyzotis, K.~Tae, and S.~Whang.
\newblock Automated data slicing for model validation: A big data - ai
  integration approach.
\newblock {\em IEEE TKDE}, 2020.

\bibitem[\protect\citeauthoryear{Cohen \bgroup \em et al.\egroup
  }{2009}]{rankdata}
Joel~W Cohen, Steven~B Cohen, and Jessica~S Banthin.
\newblock The medical expenditure panel survey: a national information resource
  to support healthcare cost research and inform policy and practice.
\newblock {\em Medical care}, pages S44--S50, 2009.

\bibitem[\protect\citeauthoryear{Commission}{1978}]{commission1978guidelines}
Equal Employment~Opportunity Commission.
\newblock Guidelines on employee selection procedures.
\newblock {\em C.F.R.}, 29., 1978.

\bibitem[\protect\citeauthoryear{Corbett-Davies \bgroup \em et al.\egroup
  }{2017}]{10.1145/3097983.3098095}
Sam Corbett-Davies, Emma Pierson, Avi Feller, Sharad Goel, and Aziz Huq.
\newblock Algorithmic decision making and the cost of fairness.
\newblock KDD, page 797–806. ACM, 2017.

\bibitem[\protect\citeauthoryear{Cortez and Silva}{2008}]{Cortez2008UsingDM}
Paulo Cortez and Alice Maria~Gon{\c{c}}alves Silva.
\newblock Using data mining to predict secondary school student performance.
\newblock EUROSIS-ETI, 2008.

\bibitem[\protect\citeauthoryear{Crenshaw}{1989}]{Crenshaw1989-CREDTI}
Kimberle Crenshaw.
\newblock Demarginalizing the intersection of race and sex: A black feminist
  critique of antidiscrimination doctrine, feminist theory and antiracist
  politics.
\newblock {\em The University of Chicago Legal Forum}, 140:139--167, 1989.

\bibitem[\protect\citeauthoryear{Dwork \bgroup \em et al.\egroup
  }{2012}]{10.1145/2090236.2090255}
Cynthia Dwork, Moritz Hardt, Toniann Pitassi, Omer Reingold, and Richard Zemel.
\newblock Fairness through awareness.
\newblock In {\em ITCS}, page 214–226. ACM, 2012.

\bibitem[\protect\citeauthoryear{Dwork}{2006}]{10.1007/11787006_1}
Cynthia Dwork.
\newblock Differential privacy.
\newblock In {\em ICALP 2006}, volume 4052 of {\em Lecture Notes in Computer
  Science}, pages 1--12. Springer Verlag, July 2006.

\bibitem[\protect\citeauthoryear{Foulds \bgroup \em et al.\egroup
  }{2018}]{Foulds2018BayesianMO}
James~R. Foulds, Rashidul Islam, Kamrun Keya, and Shimei Pan.
\newblock Bayesian modeling of intersectional fairness: The variance of bias.
\newblock In {\em SDM}, 2018.

\bibitem[\protect\citeauthoryear{Foulds \bgroup \em et al.\egroup
  }{2020}]{9101635}
James~R. Foulds, Rashidul Islam, Kamrun~Naher Keya, and Shimei Pan.
\newblock An intersectional definition of fairness.
\newblock In {\em ICDE}, pages 1918--1921, 2020.

\bibitem[\protect\citeauthoryear{Ghai \bgroup \em et al.\egroup
  }{2021}]{10.1145/3411763.3451587}
Bhavya Ghai, Md~Naimul Hoque, and Klaus Mueller.
\newblock Wordbias: An interactive visual tool for discovering intersectional
  biases encoded in word embeddings.
\newblock In {\em Extended Abstracts of CHI}, pages 1--7, 2021.

\bibitem[\protect\citeauthoryear{Ghosh \bgroup \em et al.\egroup
  }{2021}]{Ghosh2021CharacterizingIG}
A.~Ghosh, Lea Genuit, and Mary Reagan.
\newblock Characterizing intersectional group fairness with worst-case
  comparisons.
\newblock In {\em AIDBEI}, 2021.

\bibitem[\protect\citeauthoryear{Gillen \bgroup \em et al.\egroup
  }{2018}]{10.5555/3327144.3327185}
Stephen Gillen, Christopher Jung, Michael Kearns, and Aaron Roth.
\newblock Online learning with an unknown fairness metric.
\newblock In {\em NeurIPS}, page 2605–2614, 2018.

\bibitem[\protect\citeauthoryear{Gjurkovi{\'c} \bgroup \em et al.\egroup
  }{2021}]{gjurkovic-etal-2021-pandora}
Matej Gjurkovi{\'c}, Mladen Karan, Iva Vukojevi{\'c}, Mihaela Bo{\v{s}}njak,
  and Jan Snajder.
\newblock {PANDORA} talks: Personality and demographics on {R}eddit.
\newblock In {\em SocialNLP}, pages 138--152. ACL, 2021.

\bibitem[\protect\citeauthoryear{Gohar \bgroup \em et al.\egroup
  }{2022}]{gohar2022towards}
Usman Gohar, Sumon Biswas, and Hridesh Rajan.
\newblock Towards understanding fairness and its composition in ensemble
  machine learning.
\newblock {\em arXiv preprint arXiv:2212.04593}, 2022.

\bibitem[\protect\citeauthoryear{Gopalan \bgroup \em et al.\egroup
  }{2022}]{pmlr-v178-gopalan22a}
Parikshit Gopalan, Michael~P Kim, Mihir~A Singhal, and Shengjia Zhao.
\newblock Low-degree multicalibration.
\newblock In {\em COLT}, volume 178 of {\em PMLR}, pages 3193--3234, 2022.

\bibitem[\protect\citeauthoryear{Guo and
  Caliskan}{2021}]{10.1145/3461702.3462536}
Wei Guo and Aylin Caliskan.
\newblock Detecting emergent intersectional biases: Contextualized word
  embeddings contain a distribution of human-like biases.
\newblock In {\em AIES}, page 122–133. ACM, 2021.

\bibitem[\protect\citeauthoryear{Hardt \bgroup \em et al.\egroup
  }{2016}]{NIPS2016_9d268236}
Moritz Hardt, Eric Price, Eric Price, and Nati Srebro.
\newblock Equality of opportunity in supervised learning.
\newblock In {\em NeurIPS}, volume~29, 2016.

\bibitem[\protect\citeauthoryear{Harper and Konstan}{2015}]{10.1145/2827872}
F.~Maxwell Harper and Joseph~A. Konstan.
\newblock The movielens datasets: History and context.
\newblock {\em ACM Trans. Interact. Intell. Syst.}, 5(4), 2015.

\bibitem[\protect\citeauthoryear{Hashimoto \bgroup \em et al.\egroup
  }{2018}]{Hashimoto2018FairnessWD}
Tatsunori~B. Hashimoto, Megha Srivastava, Hongseok Namkoong, and Percy Liang.
\newblock Fairness without demographics in repeated loss minimization.
\newblock In {\em ICML}, 2018.

\bibitem[\protect\citeauthoryear{Hassan \bgroup \em et al.\egroup
  }{2021}]{hassan-etal-2021-unpacking-interdependent}
Saad Hassan, Matt Huenerfauth, and Cecilia~Ovesdotter Alm.
\newblock Unpacking the interdependent systems of discrimination: Ableist bias
  in {NLP} systems through an intersectional lens.
\newblock In {\em EMNLP}, pages 3116--3123. ACL, 2021.

\bibitem[\protect\citeauthoryear{Hebert-Johnson \bgroup \em et al.\egroup
  }{2018}]{pmlr-v80-hebert-johnson18a}
Ursula Hebert-Johnson, Michael Kim, Omer Reingold, and Guy Rothblum.
\newblock Multicalibration: Calibration for the
  ({C}omputationally-identifiable) masses.
\newblock In {\em ICML}, volume~80, pages 1939--1948. PMLR, 2018.

\bibitem[\protect\citeauthoryear{Huang \bgroup \em et al.\egroup
  }{2020}]{huang-etal-2020-multilingual}
Xiaolei Huang, Linzi Xing, Franck Dernoncourt, and Michael~J. Paul.
\newblock Multilingual {T}witter corpus and baselines for evaluating
  demographic bias in hate speech recognition.
\newblock In {\em LREC}, pages 1440--1448. ELRA, 2020.

\bibitem[\protect\citeauthoryear{Islam \bgroup \em et al.\egroup
  }{2019}]{islam2019mitigating}
Rashidul Islam, Kamrun~Naher Keya, Shimei Pan, and James Foulds.
\newblock Mitigating demographic biases in social media-based recommender
  systems.
\newblock {\em KDD (Social Impact Track)}, 2019.

\bibitem[\protect\citeauthoryear{Jin \bgroup \em et al.\egroup
  }{2020}]{10.1145/3318464.3384689}
Zhongjun Jin, Mengjing Xu, Chenkai Sun, Abolfazl Asudeh, and H.~V. Jagadish.
\newblock Mithracoverage: A system for investigating population bias for
  intersectional fairness.
\newblock In {\em ICDM}, page 2721–2724. ACM, 2020.

\bibitem[\protect\citeauthoryear{Kang \bgroup \em et al.\egroup
  }{2022}]{kang2022infofair}
Jian Kang, Tiankai Xie, Xintao Wu, Ross Maciejewski, and Hanghang Tong.
\newblock Infofair: Information-theoretic intersectional fairness.
\newblock In {\em IEEE Big Data}, 2022.

\bibitem[\protect\citeauthoryear{Kearns \bgroup \em et al.\egroup
  }{2018}]{pmlr-v80-kearns18a}
Michael Kearns, Seth Neel, Aaron Roth, and Zhiwei~Steven Wu.
\newblock Preventing fairness gerrymandering: Auditing and learning for
  subgroup fairness.
\newblock In {\em ICML}. PMLR, 2018.

\bibitem[\protect\citeauthoryear{Kim \bgroup \em et al.\egroup
  }{2018}]{10.5555/3327345.3327393}
Michael~P. Kim, Omer Reingold, and Guy~N. Rothblum.
\newblock Fairness through computationally-bounded awareness.
\newblock In {\em NeurIPS}, page 4847–4857, 2018.

\bibitem[\protect\citeauthoryear{Kim \bgroup \em et al.\egroup
  }{2019}]{10.1145/3306618.3314287}
Michael~P. Kim, Amirata Ghorbani, and James Zou.
\newblock Multiaccuracy: Black-box post-processing for fairness in
  classification.
\newblock In {\em AIES}, page 247–254. ACM, 2019.

\bibitem[\protect\citeauthoryear{Kiritchenko and
  Mohammad}{2018}]{kiritchenko2018examining}
Svetlana Kiritchenko and Saif Mohammad.
\newblock Examining gender and race bias in two hundred sentiment analysis
  systems.
\newblock In {\em SemEval}, pages 43--53, 2018.

\bibitem[\protect\citeauthoryear{Kirk \bgroup \em et al.\egroup
  }{2021}]{Kirk2021BiasOA}
Hannah~Rose Kirk, Yennie Jun, Haider Iqbal, Elias Benussi, Filippo Volpin,
  Fr{\'e}d{\'e}ric~A. Dreyer, Aleksandar Shtedritski, and Yuki~M. Asano.
\newblock Bias out-of-the-box: An empirical analysis of intersectional
  occupational biases in popular generative language models.
\newblock In {\em NeurIPS}, 2021.

\bibitem[\protect\citeauthoryear{Kohavi}{1996}]{Dua:2019}
Ron Kohavi.
\newblock Scaling up the accuracy of naive-bayes classifiers: A decision-tree
  hybrid.
\newblock In {\em SIGKDD}, KDD'96, page 202–207. AAAI Press, 1996.

\bibitem[\protect\citeauthoryear{Kong}{2022}]{10.1145/3531146.3533114}
Youjin Kong.
\newblock Are “intersectionally fair” ai algorithms really fair to women of
  color? a philosophical analysis.
\newblock FAccT '22, page 485–494. ACM, 2022.

\bibitem[\protect\citeauthoryear{Kusner \bgroup \em et al.\egroup
  }{2017}]{kusner2017counterfactual}
Matt~J Kusner, Joshua Loftus, Chris Russell, and Ricardo Silva.
\newblock Counterfactual fairness.
\newblock {\em In NeurIPS}, 2017.

\bibitem[\protect\citeauthoryear{Lahoti \bgroup \em et al.\egroup
  }{2020}]{NEURIPS2020_07fc15c9}
Preethi Lahoti, Alex Beutel, Jilin Chen, Kang Lee, Flavien Prost, Nithum Thain,
  Xuezhi Wang, and Ed~Chi.
\newblock Fairness without demographics through adversarially reweighted
  learning.
\newblock In {\em NeurIPS}, volume~33, pages 728--740, 2020.

\bibitem[\protect\citeauthoryear{Lalor \bgroup \em et al.\egroup
  }{2022}]{lalor-etal-2022-benchmarking}
John Lalor, Yi~Yang, Kendall Smith, Nicole Forsgren, and Ahmed Abbasi.
\newblock Benchmarking intersectional biases in {NLP}.
\newblock In {\em NAACL}, pages 3598--3609. ACL, 2022.

\bibitem[\protect\citeauthoryear{Liu \bgroup \em et al.\egroup
  }{2015}]{liu2018large}
Ziwei Liu, Ping Luo, Xiaogang Wang, and Xiaoou Tang.
\newblock Deep learning face attributes in the wild.
\newblock In {\em ICCV}, December 2015.

\bibitem[\protect\citeauthoryear{Martinez \bgroup \em et al.\egroup
  }{2021}]{pmlr-v139-martinez21a}
Natalia~L Martinez, Martin~A Bertran, Afroditi Papadaki, Miguel Rodrigues, and
  Guillermo Sapiro.
\newblock Blind pareto fairness and subgroup robustness.
\newblock In {\em ICML}, PMLR, pages 7492--7501, 2021.

\bibitem[\protect\citeauthoryear{Mas-Colell \bgroup \em et al.\egroup
  }{1995}]{mas-colell.whinston.ea95}
Andreu Mas-Colell, Michael~D. Whinston, and Jerry~R. Green.
\newblock {\em Microeconomic Theory}.
\newblock Oxford University Press, New York, 1995.

\bibitem[\protect\citeauthoryear{May \bgroup \em et al.\egroup
  }{2019}]{may2019measuring}
Chandler May, Alex Wang, Shikha Bordia, Samuel Bowman, and Rachel Rudinger.
\newblock On measuring social biases in sentence encoders.
\newblock In {\em 2019 NAACL}, 2019.

\bibitem[\protect\citeauthoryear{Mehrabi \bgroup \em et al.\egroup
  }{2021}]{10.1145/3457607}
Ninareh Mehrabi, Fred Morstatter, Nripsuta Saxena, Kristina Lerman, and Aram
  Galstyan.
\newblock A survey on bias and fairness in machine learning.
\newblock {\em ACM Comput. Surv.}, 54(6), 2021.

\bibitem[\protect\citeauthoryear{Molina and Loiseau}{2022}]{molina2022bounding}
Mathieu Molina and Patrick Loiseau.
\newblock Bounding and approximating intersectional fairness through marginal
  fairness.
\newblock {\em arXiv preprint arXiv:2206.05828}, 2022.

\bibitem[\protect\citeauthoryear{Morina \bgroup \em et al.\egroup
  }{2019}]{DBLP:journals/corr/abs-1911-01468}
Giulio Morina, Viktoriia Oliinyk, Julian Waton, Ines Marusic, and Konstantinos
  Georgatzis.
\newblock Auditing and achieving intersectional fairness in classification
  problems.
\newblock {\em CoRR}, abs/1911.01468, 2019.

\bibitem[\protect\citeauthoryear{Rawls}{2001}]{rawls2001justice}
John Rawls.
\newblock {\em Justice as fairness: {A} restatement}.
\newblock Harvard University Press, 2001.

\bibitem[\protect\citeauthoryear{Shannon}{1948}]{shannon1948mathematical}
Claude~E Shannon.
\newblock A mathematical theory of communication.
\newblock {\em The Bell system technical journal}, 27(3):379--423, 1948.

\bibitem[\protect\citeauthoryear{Shui \bgroup \em et al.\egroup
  }{2022}]{shui2022on}
Changjian Shui, Gezheng Xu, Qi~CHEN, Jiaqi Li, Charles Ling, Tal Arbel, Boyu
  Wang, and Christian Gagn{\'e}.
\newblock On learning fairness and accuracy on multiple subgroups.
\newblock In {\em NeurIPS}, 2022.

\bibitem[\protect\citeauthoryear{Steed and
  Caliskan}{2021}]{10.1145/3442188.3445932}
Ryan Steed and Aylin Caliskan.
\newblock Image representations learned with unsupervised pre-training contain
  human-like biases.
\newblock FAccT '21, page 701–713. ACM, 2021.

\bibitem[\protect\citeauthoryear{Subramanian \bgroup \em et al.\egroup
  }{2021}]{subramanian-etal-2021-evaluating}
Shivashankar Subramanian, Xudong Han, Timothy Baldwin, Trevor Cohn, and Lea
  Frermann.
\newblock Evaluating debiasing techniques for intersectional biases.
\newblock In {\em EMNLP}, pages 2492--2498. ACL, 2021.

\bibitem[\protect\citeauthoryear{Tan and Celis}{2019}]{10.5555/3454287.3455472}
Yi~Chern Tan and L~Elisa Celis.
\newblock Assessing social and intersectional biases in contextualized word
  representations.
\newblock {\em In NeurIPS}, 32, 2019.

\bibitem[\protect\citeauthoryear{Wang \bgroup \em et al.\egroup
  }{2022}]{10.1145/3531146.3533101}
Angelina Wang, Vikram~V Ramaswamy, and Olga Russakovsky.
\newblock Towards intersectionality in machine learning: Including more
  identities, handling underrepresentation, and performing evaluation.
\newblock In {\em FAccT}, page 336–349. ACM, 2022.

\bibitem[\protect\citeauthoryear{Wightman}{1998}]{Wightman1998LSACNL}
Linda~F. Wightman.
\newblock {\em LSAC National Longitudinal Bar Passage Study}.
\newblock LSAC research report series. Law School Admission Council, 1998.

\bibitem[\protect\citeauthoryear{Yang \bgroup \em et al.\egroup
  }{2020a}]{NEURIPS2020_29c0605a}
Forest Yang, Mouhamadou Cisse, and Sanmi Koyejo.
\newblock Fairness with overlapping groups; a probabilistic perspective.
\newblock In {\em NeurIPS}, volume~33, pages 4067--4078, 2020.

\bibitem[\protect\citeauthoryear{Yang \bgroup \em et al.\egroup
  }{2020b}]{yang2021causal}
Ke~Yang, Joshua~R Loftus, and Julia Stoyanovich.
\newblock Causal intersectionality for fair ranking.
\newblock 2020.

\bibitem[\protect\citeauthoryear{Yona and Rothblum}{2018}]{yona2018probably}
Gal Yona and Guy Rothblum.
\newblock Probably approximately metric-fair learning.
\newblock In {\em ICML}, pages 5680--5688. PMLR, 2018.

\bibitem[\protect\citeauthoryear{Zhang \bgroup \em et al.\egroup
  }{2017}]{zhifei2017cvpr}
Zhifei Zhang, Yang Song, and Hairong Qi.
\newblock Age progression/regression by conditional adversarial autoencoder.
\newblock In {\em CVPR}, pages 5810--5818, 2017.

\end{thebibliography}

\end{document}